\documentclass{article}
\usepackage{float}
\usepackage{acra}
\usepackage{flafter}
\usepackage{graphicx}
\usepackage{listings}
\usepackage{url}
\usepackage{multirow}

\title{Multi-Modal Camera-Based Detection of Vulnerable Road Users}
\author{Penelope Brown, Julie Stephany Berrio Perez, Mao Shan, Stewart Worrall \\ The University of Sydney, Australian Centre for Robotics \\ 
pbro3848@uni.sydney.edu.au, \{j.berrio, m.shan, s.worrall\}@acfr.usyd.edu.au
\thanks{This research was supported by the 2025 Engineering Vacation Research Internship (VRI) Winter Program.}}

\begin{document}
\maketitle

\pagestyle{plain}
\pagenumbering{arabic}

\begin{abstract}
\textbf{Vulnerable road users (VRUs) such as pedestrians, cyclists, and motorcyclists represent more than half of global traffic deaths, yet their detection remains challenging in poor lighting, adverse weather, and unbalanced data sets. This paper presents a multimodal detection framework that integrates RGB and thermal infrared imaging with a fine-tuned YOLOv8 model. Training leveraged KITTI, BDD100K, and Teledyne FLIR datasets, with class re-weighting and light augmentations to improve minority-class performance and robustness, experiments show that 640-pixel resolution and partial backbone freezing optimise accuracy and efficiency, while class-weighted losses enhance recall for rare VRUs. Results highlight that thermal models achieve the highest precision, and RGB-to-thermal augmentation boosts recall, demonstrating the potential of multimodal detection to improve VRU safety at intersections. }
\end{abstract}

\section{Introduction}



\begin{figure}[t]
    \centering
    \includegraphics[trim={0 7.8cm 0 0},clip, width=0.9\linewidth]{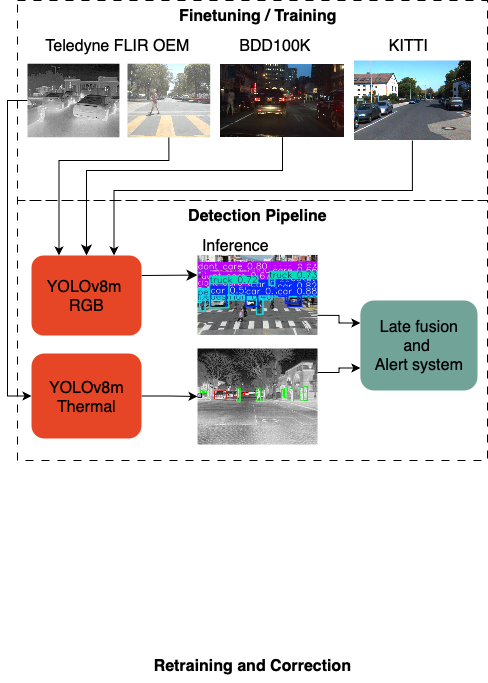}
    \caption{\small Overview of the multimodal training pipeline integrating RGB and thermal data for vulnerable road user detection using the YOLOv8 model}
    \label{fig:placeholder}
\end{figure}

Road safety is a major concern throughout the world, particularly in relation to protecting vulnerable road users (VRUs). Globally, road traffic accidents claim about 1.2 million lives per year, and more than half of these fatalities are VRUs \cite{who}. Many regions are seeing increasing trends in VRU casualties despite overall safety efforts. For example, Australia recorded 1,272 road deaths in 2023 (a 7.3\% increase from 2022) \cite{annual_trauma}, with deaths among VRUs showing the highest surge of all user groups since 2020. VRUs are broadly defined to include pedestrians, cyclists, motorcyclists, riders of micromobility devices (such as e-scooters), users of mobility aids (e.g. wheelchairs) \cite{Fact_sheet_vru}, and road crossing animals. Improving the safety of these diverse participants is a complex problem, as VRUs are typically small, move at inconsistent speeds, and exhibit unpredictable behaviour in traffic \cite{4}. Their lack of physical protection makes them disproportionately vulnerable in collisions and accounts for an acute and growing share of road fatalities, underscoring the need for effective detection and warning systems.

Advances in computer vision and sensor technology offer new opportunities to mitigate VRU risks through early detection and alerting. In advanced driver assistance systems (ADAS) and intelligent transportation infrastructure, the ability to automatically detect VRUs in real time is needed to prevent accidents or trigger timely warnings. However, reliable VRU detection poses several key challenges. First, many VRUs (especially pedestrians and cyclists) appear as relatively small objects in images or sensor data, making them harder to detect than larger vehicles. Second, VRU detection is inherently a multiclass problem, the system must recognise a variety of classes (people, bikes, motorcycles, animals, etc.), which complicates training and often leads to unbalanced performance.

A further challenge lies in environmental and sensor limitations. Most VRU detection implementations to date rely on a single sensor modality, often a visible-light (RGB) camera, due to its rich image detail and low cost. However, relying solely on an optical camera can be problematic in conditions such as night, glare, or fog, when visible-spectrum imaging may fail to clearly capture pedestrians or cyclists. Thermal infrared cameras, on the other hand, passively sense heat emissions and can readily highlight humans or animals in darkness or high-contrast lighting, but thermal imagery lacks colour/texture details and may have lower resolution. No single sensor is universally reliable in all scenarios. This limitation has prompted a trend toward multi-modal approaches in recent VRU detection research. 

Currently, the rapid progress of the past decade in object detection algorithms has equipped us with powerful models capable of accurately recognising various road users in real time.  The introduction of the original YOLO model in 2016 demonstrated that a single CNN could directly predict bounding boxes and class probabilities at 45 frames per second. YOLOv8 (released in 2023) incorporates advanced network designs and training strategies that offer enhanced performance and robustness compared to previous versions. These state-of-the-art detectors provide a strong foundation for tackling the VRU detection problem, as they can be fine-tuned on specialised datasets to recognise the critical VRU classes. However, achieving an equally reliable detection for VRUs classes in challenging low-light conditions still requires careful consideration of training methodologies, data enhancement, and sensor input.

In this paper, we address the above challenges by proposing a multimodal VRU detection approach that uses RGB and thermal camera data with a cutting-edge object detector. The system is designed for deployment at road intersections to actively identify nearby VRUs (e.g. pedestrians, cyclists, and scooter riders) and alert drivers in real time. We adopt a fine-tuned YOLOv8 model as the core detector, taking advantage of its balance of speed and accuracy for real-time performance. To boost detection of under-represented VRU classes, we employ targeted training enhancements and extensive data augmentation to improve diversity. These techniques are vital for developing a robust high-performance model that is suited to the complexities of VRU detection. In the following sections, we detail the related work in this domain and then present our methodology and results.

\section{Related Work}

Early object detection approaches combined innovative features with boosting and cascades to achieve real-time performance. Viola and Jones \cite{13} pioneered a face detector using an \textit{Integral Image} for a rapid Haar-like feature computation, enabling AdaBoost-based detection at 15 frames per second. Shortly after, Dalal and Triggs \cite{Dalal2005} introduced Histograms of Orientated Gradients (HOG) within a sliding window framework, attaining near-perfect pedestrian detection in the MIT dataset and establishing the efficacy of robust feature descriptors for object detection. The rise of deep learning further revolutionised the field: He \emph{et al.} \cite{He2016} proposed the deep Residual Network (ResNet) that won the 2015 ImageNet competition by enabling substantially increased network depth without vanishing gradients, marking a new state of the art in image feature learning.

Modern detectors build on these foundations with region proposal mechanisms and end-to-end training. Girshick \emph{et al.} \cite{Girshick2016} introduced the Region-Based Convolutional Neural Network (R-CNN), which improved the mean average precision (mAP) by more than 50\% relative to the previous PASCAL VOC champion, although at the cost of multiple-stage processing. Ren \emph{et al.} \cite{Ren2017} later developed Faster R-CNN, integrating a learnable region proposal network for near real-time performance while maintaining high detection accuracy. In parallel, Redmon \emph{et al.} \cite{Redmon2016} pioneered one-stage detectors with the You Only Look Once (YOLO) model, which forwent explicit proposal generation in favour of a single unified network, dramatically increasing the inference speed to 45~FPS with a slight trade-off in localisation precision. Subsequent YOLO variants (v2 through v8) progressively improved the accuracy-speed trade-off, incorporating advances like multiscale predictions and deeper backbones. The latest YOLOv8, introduced in 2023 \cite{Varghese2024,UltralyticsY8}, achieves state-of-the-art real-time performance on general object benchmarks. Although even more recent iterations (e.g., a YOLOv11 prototype) report further accuracy gains, they demand greater computational power and lack the ecosystem maturity of YOLOv8 \cite{UltralyticsComparison}. Consequently, YOLOv8 remains a popular choice for real-time detection tasks, offering a robust baseline upon which our work builds on for vulnerable road user recognition.

Despite progress in generic object detection, detecting vulnerable road users (VRUs), such as pedestrians and cyclists, presents ongoing challenges. VRU instances are often small and occur in diverse forms, causing their detection benchmarks to lag behind those of common objects. For example, Zhou and Tuzel \cite{Zhou2017VoxelNet} proposed \textit{VoxelNet}, a LiDAR-based 3D detection network with a region proposal stage, which achieved only moderate success on the KITTI benchmark (approximately 58\% and 67\% AP for pedestrians and cyclists, respectively, under easy conditions). On the 2D vision side, Castello \emph{et al.} \cite{Castello2020} showed that modifying models to driving datasets improves performance: they fine-tuned YOLOv3/v4 on the BDD100K dataset and attained mAP scores of 47---- 64\% for person, bicycle and motorcycle classes, significantly outperforming the original MS-COCO training of those models in the same categories. Similarly, Narkhede and Chopade \cite{Narkhede2024} reported a high $0.76$ average mAP in the classes of pedestrians, cyclists and motorcyclists after fine-tuning the YOLOv8 detector on a custom VRU image set, underscoring the value of the specialisation of the data set. Most recently, Liu and Shi \cite{Liu2025} introduced \textit{VRU-YOLO}, a YOLO-based small object detector optimised for complex traffic scenes, further reflecting the trend toward specialised architectures for VRU detection.

Beyond the main VRU groups, several works address rare or emerging VRU categories that standard detectors struggle with. For example, e-scooter riders have a unique appearance and dynamic behaviour, prompting Gilroy \emph{et al.} \cite{Gilroy2022} to develop a dedicated benchmark and detection model for partially occluded e-scooter users. Their approach improved the accuracy of the rider classification by about 15. 9\% compared to a prior solution based on YOLOv3 + MobileNet (Apurv \emph{et al.} 2021). In a different niche, Beyer \emph{et al.} (2016) designed the DROW detector, a 2D convolutional model operating on laser range data, to specifically recognise wheelchairs in elderly care environments. Likewise, animal-vehicle collisions have motivated wildlife detection systems using YOLO variants: studies have demonstrated detectors for region-specific animals (e.g., detecting deer or livestock on roads) using deep learning models \cite{Roopashree2021,Dave2023,Rajesh2024}. However, these efforts towards rare-class VRU detection remain fragmented. Methods are often tailored to a single category and evaluated on disparate datasets, making direct comparison difficult. There is a clear gap in developing a unified detection framework that robustly handles *all* VRU types, including those with limited training examples. Our work addresses this gap by employing targeted training techniques (such as class re-weighting and data augmentation) to boost the detection of under-represented VRU classes, while maintaining strong performance on common VRUs.

To mitigate the limitations of any individual sensor \cite{9406432}, recent VRU detection research has leaned toward multimodal sensor fusion \cite{Ahmar2023, Zhang2025Survey}. Combining data from cameras, lasers, and other sensors can substantially improve detection robustness; for example, Wan \emph{et al.} introduced a multisensor platform incorporating RGB cameras, thermal infrared imagers, LiDAR, and radar, and found that adding a thermal camera was essential to reliably detect VRUs in darkness or glare \cite{Wan2018}. However, traditional optical cameras remain the most ubiquitous sensing modality for VRU detection due to their high resolution and low cost \cite{Zhang2025Survey}. This has driven interest in enhancing camera-based systems by fusing them with complementary spectra like thermal infrared (far-IR). Thermal imagery highlights the heat signatures of pedestrians and cyclists, offering resilience in low-light or high-contrast scenarios where RGB alone may fail.

\section{Methodology}

This study develops a multi-modal VRU detection framework that combines RGB and thermal imaging with a fine-tuned YOLOv8 model. The methodology focuses on dataset preparation, augmentation, backbone freezing, and class re-weighting to optimsze performance.

\subsection{System Setup}

\textbf{Sensors:} The proposed detection framework employs a dual camera setup consisting of a standard RGB camera and a thermal infrared (TIR) camera. RGB cameras capture rich colour and texture information under normal lighting, while thermal cameras capture emitted infrared radiation, which highlights warm-bodied VRUs such as pedestrians, even in poor visibility scenarios. 

\textbf{Detector:} The detection backbone is based on a fine-tuned YOLOv8 model, chosen for its balance of detection accuracy and high frame rate inference, making it well suited for real-time roadside applications.

\subsection{Datasets}


In our system, a YOLOv8 detection model is selected as the core backbone for VRU recognition due to its balance of accuracy and real-time inference performance. The model is fine-tuned in multiple open-source autonomous driving datasets, including the KITTI 2D Object dataset \cite{Geiger2013IJRR}, the BDD100K dataset \cite{bdd100k}, and the Teledyne FLIR OEM RGB paired dataset \cite{teledyne_flir}. This composite training strategy increases data diversity, exposes the model to a broader spectrum of environmental conditions and VRU appearances, and helps mitigate overfitting to any single-sensor configuration. Using both visible-spectrum and thermal imagery during training, the model gains robustness across varied illumination, weather, and occlusion scenarios.

\begin{table}[h]
    \centering
    \begin{tabular}{|p{4.7em}|p{4.7em}|p{4.7em}|p{4.7em}|}
    \hline
        \textbf{\small Final Class Set} & \textbf{\small KITTI} & \textbf{\small BDD100K} & \textbf{\small Teledyne FLIR} \\
        \hline
        \hline
        \small Car & \small Car, Van & \small Car & \small Car \\
        \hline
        \small Pedestrian & \small Pedestrian, Person\_sitting &\small Person, rider & \small Person, people, stroller \\
        \hline
        \small Cyclist & \small Cyclist & \small Bike & \small Bike \\
        \hline
        \small Bus &\small Bus &\small Bus &\small Bus \\
        \hline
        \small Truck & Truck & truck & truck \\
        \hline
        \small Animal & \small Animal & \small N/A & \small Dog \\
        \hline
        \small Motorcycle & \small Motorcycle & \small Motor & \small Motor \\
        \hline
        \small Scooter & \small Scooter & \small N/A & \small Scooter \\
        \hline
        \small Other vehicle & \small Tram, Misc & \small Train & \small Train, other vehicle \\
        \hline
        \small Don't care & \small Don't care & \small Traffig sign, traffic light & \small Skateboard, light, hydrant, sign \\
        \hline
    \end{tabular}
    \caption{\small Custom Dataset Mapping}
    \label{Custom Dataset Mapping}
\end{table}

Publicly available datasets provide essential benchmarks for developing and evaluating object detection models, as they enable both training diversity and direct performance comparison with prior work. The KITTI 2D Object dataset comprises 7,481 training RGB images and 7,518 test RGB images, making it one of the most widely used benchmarks in autonomous driving research \cite{Geiger2013IJRR}. The Berkeley DeepDrive (BDD100K) dataset offers significantly greater scale, containing 100,000 annotated RGB images divided into 70,000 training, 10,000 validation, and 20,000 test samples \cite{bdd100k}. To incorporate thermal imagery and support multimodal learning, the Teledyne FLIR data set is also used, consisting of 10,319 RGB and 10,742 thermal training images, 1,085 RGB and 1,144 thermal validation images, and 3,749 RGB–thermal test pairs \cite{teledyne_flir}.

\begin{table}[h]
    \centering
    \begin{tabular}{|p{4em}|p{3em}|p{3em}|p{4em}|p{4em}|p{4em}|}
    \hline
         \small \textbf{Class} & \small \textbf{RGB Train} & \small \textbf{RGB Val} & \small \textbf{Thermal Train} & \small \textbf{Thermal Val} \\
         \hline
         \hline
         \small Car & 817058 & 109825 & 73623 & 7133 \\
         \hline
         \small Pedestrian & 129825 & 16985 & 44542 & 4315 \\
         \hline
         \small Cyclist & 16414 & 1200 & 7237 & 170 \\
         \hline
         \small Truck & 32357 & 4294 & 829 & 46 \\
         \hline
         \small Bus & 13567 & 1780 & 2245 & 179 \\
         \hline
         \small Animal & 0 & 0 & 4 & 0 \\
         \hline
         \small Motor cycle & 4839 & 529 & 1116 & 66 \\
         \hline
         \small Scooter & 41 & 0 & 15 & 0 \\
         \hline
         \small Other Vehicle & 2327 & 55 & 1378 & 63 \\
         \hline
         \small  Don't Care & 487128 & 67660 & 38092 & 4574 \\
         \hline
    \end{tabular}
    \caption{\small Class instances in combined dataset}
    \label{Class Instances in Combined Dataset}
\end{table}

To unify these heterogeneous sources, a custom label mapping strategy was designed (Table~\ref{Custom Dataset Mapping}), ensuring consistent class definitions in the datasets. After preprocessing in the YOLOv8 format, the resulting data set statistics are summarised in Table~\ref{Combined Dataset}. It is important to note that discrepancies between image and label counts arise from the FLIR dataset, which excludes annotations for images without objects of interest. Finally, Table~\ref{Class Instances in Combined Dataset} reports the class-wise distribution of instances, highlighting imbalances that motivate the use of customised training strategies to enhance performance in under-represented VRU categories.



\begin{table}[h]
    \centering
    \begin{tabular}{|l|l|l|}
    \hline
        \textbf{Dataset} & \textbf{Images} & \textbf{Labels} \\
        \hline
        \hline
        RGB Train & 87800 & 87534 \\
        \hline
        RGB Val & 11085 & 11069 \\
        \hline
        Thermal Train & 10742 & 
        10478 \\
        \hline
        Thermal Val & 1144 & 1128 \\
        \hline
    \end{tabular}
    \caption{\small Combined dataset statistics}
    \label{Combined Dataset}
\end{table}

Once deployed on-site, the two models operate within a late-fusion framework, where each independently processes incoming frames, and their outputs are subsequently combined. The final detection confidence is derived by weighting the relative confidence scores of the individual models, thereby leveraging the complementary strengths of both modalities.

\subsection{Dataset Augmentation}

A series of experiments were conducted on the combined datasets to identify training parameters and to evaluate the effects of image filtering and enhancement strategies on model performance; these experiments focused on improving robustness.

\subsubsection{Image Resolution}

We examined the impact of input image resolution on detection accuracy and computational efficiency. Although lower resolutions reduce memory usage and inference time, an important factor for edge-based deployment, they may also lead to substantial performance degradation due to the loss of fine-grained visual features. To investigate this trade-off, a subset of 500 RGB images from the combined training set was used to train the YOLOv8m model for 50 epochs. During training, ten layers of the backbone were frozen to preserve pre-trained feature representations, and a batch size of four was employed. 

\subsubsection{Image Translation}

To further enhance the robustness of the model in diverse conditions, we evaluated augmentation strategies using the \texttt{albumentationslibrary}. Two augmentation regimes were tested. The \textit{heavy} augmentation pipeline included transformations such as \texttt{CoarseDropout} or \texttt{GridDropout}, \texttt{ToGray} or \texttt{ChannelDropout}, \texttt{RandomBrightnessContrast}, \texttt{GaussianBlur} or \texttt{GaussNoise}, and weather-based effects including \texttt{RandomRain}, \texttt{RandomFog}, or \texttt{RandomSnow}. In contrast, the \textit{light} augmentation pipeline applied a reduced set of transformations consisting of \texttt{RandomBrightnessContrast}, \texttt{GaussianBlur}, \texttt{GaussNoise}, and one weather effect (\texttt{RandomRain}, \texttt{RandomFog}, or \texttt{RandomSnow}). These strategies were designed to simulate challenging real-world conditions and improve model generalisation.

\begin{figure}[h]
    \centering
    \includegraphics[width=1.0\linewidth]{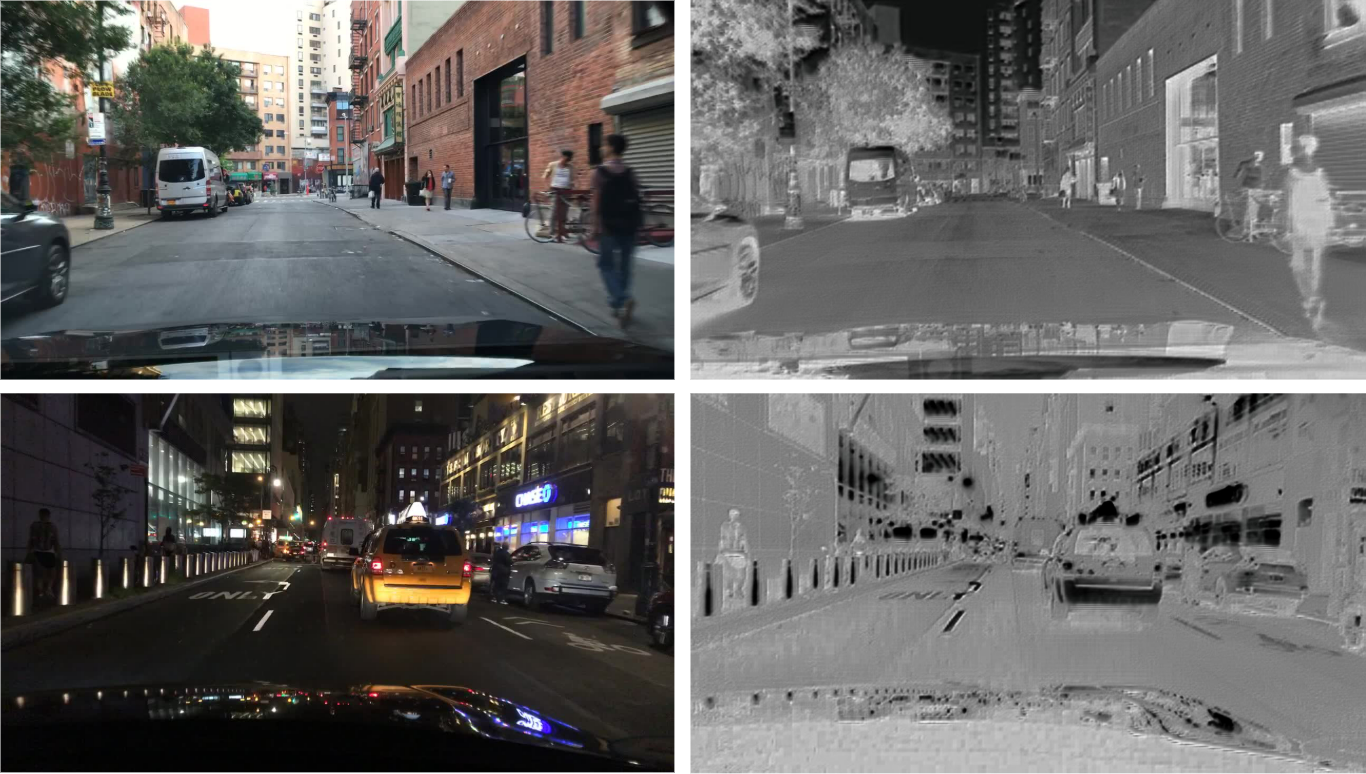}
    \caption{\small Examples of RGB images converted to thermal infrared (TIR) images via \cite{lee2023edge} image translation technique to enhance thermal dataset diversity.}
    \label{RGB to TIR Image Conversion}
\end{figure}

In addition, the thermal data set was expanded using a novel RGB-to-thermal \cite{lee2023edge} image conversion technique, some examples can be seen in Fig. \ref{RGB to TIR Image Conversion}. This image-to-image translation approach enabled the generation of synthetic thermal images from the RGB data set, thus increasing the diversity of the thermal training set. The inclusion of such translated images was designed to strengthen the model’s ability to detect VRUs in thermal images, particularly under conditions where real thermal data are limited.

\subsection{Training Methodology}

\subsubsection{Backbone Freezing}

Fine-tuning the YOLOv8m model for custom datasets provides a computationally efficient strategy for adapting pre-trained networks to specialised tasks. This process typically involves \textit{freezing} a portion of the backbone layers, thereby preserving low-level feature representations while retraining the neck and head layers to learn task-specific classes. By limiting the number of trainable parameters, backbone freezing reduces both training time and computational cost. However, excessive freezing can be detrimental, particularly in multiclass detection tasks, as it restricts the model's ability to adapt higher-level features to novel classes. Thus, selecting an appropriate balance in the number of frozen layers is critical to achieving strong performance in various VRU categories.

\subsubsection{Class Imbalance}
Training directly in the full combined dataset introduces challenges due to a significant \textit{class imbalance}, as illustrated in Table~\ref{Class Instances in Combined Dataset}. For example, common classes such as cars dominate the dataset, whereas minority VRU classes (e.g., scooters or motorcycles) are under-represented. Without correction, this imbalance biases the model toward majority classes, resulting in poor recall for rare but safety-critical VRU categories. 

To address this issue, two complementary strategies were investigated. The first strategy involved \textit{selective dataset filtration}, whereby under-represented or ambiguous classes were emphasised while less relevant categories were pruned, allowing the model to focus on VRUs. The second strategy modified the YOLOv8m loss function to apply \textit{class-weighted penalties}: misclassifications of minority classes incurred heavier penalties than errors in majority classes. This adjustment forced the training process to allocate greater learning capacity to rare VRU categories, thus enhancing their detection accuracy without severely compromising performance in dominant classes.

\section{Results}

This section presents the analysis of experiments conducted to evaluate image resolution, backbone freezing, class imbalance strategies, augmentation methods, and final model performance.

\subsection{Impact of Image Resolution}

The experimental comparison between an input size of 320 pixels and 640 pixels is summarised in Table~\ref{Image Resizing Experiment Results}. As shown, the 640-pixel configuration achieved a substantial improvement in detection metrics, with an increase in mAP50 from 0.303 to 0.459 and in mAP50:90 from 0.176 to 0.261. Similarly, improvements in the regression of the boundary box (box) and the recall (R) were observed, with values rising from 0.42 to 0.607 and 0.313 to 0.416, respectively. 

\begin{table}[h]
\centering
\begin{tabular}{|c|c|c|c|c|}
\hline
\textbf{\small Resolution} & \textbf{\small mAP50} & \small \textbf{mAP50:90} & \small \textbf{Box} & \small \textbf{R} \\
\hline
\hline
320 & 0.303 & 0.176 & 0.42  & 0.313 \\
\hline
640 & 0.459 & 0.261 & 0.607 & 0.416 \\
\hline
\end{tabular}
\caption{\small Image Resizing Experiment Results}
\label{Image Resizing Experiment Results}
\end{table}

These results demonstrate that, while higher resolution increases computational cost, it provides a clear accuracy advantage that is essential for small-object VRU detection. In particular, the performance at 320 pixels was insufficient for reliable deployment, whereas the 640-pixel input yielded significantly stronger detection capabilities, justifying its selection as the optimal resolution for subsequent experiments.

\subsection{Backbone Freezing Results}

To evaluate the impact of backbone freezing on detection performance, we conducted an experiment using the same parameters as the image resolution study, with an input image size of 640 pixels. Starting from a coco-pretrained model, two configurations were compared: freezing six backbone layers versus freezing the entire backbone (ten layers). The results are presented in Table~\ref{Layer Freezing Experiment Results}.

\begin{table}[H]
    \centering
    \begin{tabular}{|c|c|c|c|c|}
    \hline
        \multicolumn{1}{|c|}{\textbf{\begin{tabular}[c]{@{}c@{}}Frozen\\ layers\end{tabular}}} & \textbf{mAP50} & \textbf{mAP50:90} & \textbf{Box} & \textbf{R} \\
        \hline \hline
        6  & 0.481 & 0.278 & 0.610 & 0.440 \\ \hline
        10 & 0.459 & 0.261 & 0.607 & 0.316 \\
        \hline
    \end{tabular}
    \caption{Layer Freezing Experiment Results}
    \label{Layer Freezing Experiment Results}
\end{table}

As shown in Table~\ref{Layer Freezing Experiment Results}, freezing six layers of the backbone resulted in better performance in all metrics evaluated compared to freezing all ten layers. Specifically, the six-layer configuration achieved higher mAP50 (0.481 vs. 0.459), mAP50:90 (0.278 vs. 0.261), bounding box regression accuracy (0.610 vs. 0.607) and recall (0.440 vs. 0.316). These findings confirm that while partial backbone freezing can effectively reduce training cost and preserve low-level feature representations, excessive freezing restricts the model’s ability to adapt to novel VRU classes, leading to degraded multiclass detection performance. 

\subsection{Class Imbalance Results}

The experiments on class imbalance investigated the effect of dataset filtration and loss function modification on detection performance. The combined dataset is dominated by majority classes, such as cars, while critical VRU categories (e.g., motorcycles and scooters) are sparsely represented. Without corrective measures, this imbalance biases the detector toward frequent classes at the expense of rare but safety-relevant VRUs.  

To mitigate the influence of dominant classes, we first examined the effect of selective filtration of data sets. Three configurations were compared: the original full dataset, a 4-Class subset focussing exclusively on VRUs (pedestrian, cyclist, motorcycle, scooter), and a 7-Class subset that retained key VRU categories alongside representative vehicle classes. All models were trained on a 4,000-image RGB subset with an input size of 640 pixels, six frozen backbone layers, and a batch size of four.  

\begin{table}[h]
    \centering
    \begin{tabular}{|c|cc|cc|}
\hline
\multicolumn{1}{|c|}{\multirow{2}{*}{\textbf{Class}}} & \multicolumn{2}{c|}{\textbf{7-class model}}                             & \multicolumn{2}{c|}{\textbf{4-class model}}                             \\ \cline{2-5} 
\multicolumn{1}{|c|}{}                                & \multicolumn{1}{c|}{\textbf{Train}} & \multicolumn{1}{c|}{\textbf{Val}} & \multicolumn{1}{c|}{\textbf{Train}} & \multicolumn{1}{c|}{\textbf{Val}} \\ \hline
          \hline
          \hline
         Car & \multicolumn{1}{c|}{31263} & 7668 & \multicolumn{1}{c|}{0} & 0\\
          \hline
         Pedestrian & \multicolumn{1}{c|}{4880} & 1140 & \multicolumn{1}{c|}{4880} & 1140\\
          \hline
         Cyclist & \multicolumn{1}{c|}{580} & 156& \multicolumn{1}{c|}{580} & 156 \\
          \hline
         Truck & \multicolumn{1}{c|}{1297} & 271& \multicolumn{1}{c|}{0} & 0 \\
          \hline
         Bus & \multicolumn{1}{c|}{515} & 111& \multicolumn{1}{c|}{0} & 0 \\
          \hline
         Motorcycle & \multicolumn{1}{c|}{176} & 33& \multicolumn{1}{c|}{176} & 33 \\
          \hline
         Scooter & \multicolumn{1}{c|}{3} & 0 & \multicolumn{1}{c|}{3} & 0 \\
          \hline
    \end{tabular}
    \caption{\small Class instances for 7-class and 4-class models}
    \label{Class Instances for 7-Class Model}
\end{table}

The results are presented in Table~\ref{Results for Class Filtration Experiment}. The 7-Class configuration achieved the highest overall performance, with an mAP50 of 0.517 and recall of 0.462, outperforming both the full dataset and the 4-Class subset. By contrast, the 4-Class model demonstrated significantly reduced performance (mAP50 of 0.273), which can be attributed to the severe shortage of training samples after excessive class removal. Table ~\ref{Class Instances for 7-Class Model} illustrates the reduction in class instances, highlighting that aggressive filtration led to insufficient data for effective learning. These findings suggest that moderate filtration, which reduces noise without eliminating too much data, is more beneficial for VRU detection.



\begin{table}[h]
    \centering
    \begin{tabular}{|c|c|c|c|c|}
    \hline
        \small \textbf{Model} & \small \textbf{mAP50} & \small \textbf{mAP50:90} & \small \textbf{Box} & \small \textbf{Recall} \\
        \hline
        \hline
        Full Class & 0.481 & 0.278 & 0.610 & 0.440 \\
        \hline
        4-Class & 0.273 & 0.132 & 0.448 & 0.258 \\
        \hline
        7-Class & 0.517 & 0.300 & 0.663 & 0.462 \\
        \hline
    \end{tabular}
    \caption{\small Results for class filtration experiment}
    \label{Results for Class Filtration Experiment}
\end{table}

 The second strategy involved modifying the YOLOv8m loss function to incorporate class-weighted penalties, thus prioritising minority class detections. As shown in Table~\ref{Adjusted Loss Function Experiment Results}, this adjustment yielded substantial improvements to the 7-class subset. Specifically, recall increased from 0.300 to 0.480, and the bounding-box regression improved from 0.517 to 0.646. Although the mAP50:90 score decreased slightly (0.462 to 0.316), the overall gains in recall and minority class sensitivity justify the trade-off, particularly for safety-critical VRU applications.  

\begin{table}[H]
    \centering
    \begin{tabular}{|p{5em}|c|c|c|c|}
    \hline
        \small \textbf{Model} & \small \textbf{Box} & \small \textbf{Recall} & \small \textbf{mAP50} & \small \textbf{mAP50:90} \\
        \hline
        \hline
        7-Class & 0.517 & 0.300 & 0.663 & 0.462 \\
        \hline
        Adjusted 7-Class & 0.646 & 0.480 & 0.531 & 0.316 \\
        \hline
    \end{tabular}
    \caption{\small Adjusted loss function experiment results}
    \label{Adjusted Loss Function Experiment Results}
\end{table}

\begin{table}[H]
    \centering
    \begin{tabular}{|p{5em}|c|c|c|c|}
    \hline
       \small \textbf{Model} &\small  \textbf{Box} &\small  \textbf{Recall} &\small  \textbf{mAP50} &\small  \textbf{mAP50:90} \\
        \hline
        \hline
        Original Thermal & 0.404 & 0.287 & 0.287 & 0.147 \\
        \hline
        Modified Thermal & 0.780 & 0.638 & 0.724 & 0.442 \\
        \hline
    \end{tabular}
    \caption{\small Original vs modified thermal dataset performance}
    \label{Original vs Modified Thermal Dataset Performance}
\end{table}

To assess the generality of the class-weighted loss strategy, the same modification was applied to the full thermal dataset. Table~\ref{Original vs Modified Thermal Dataset Performance} shows a dramatic performance improvement: mAP50 increased from 0.287 to 0.724, and recall increased more than doubled from 0.287 to 0.638. These results confirm the effectiveness of class-weighted penalties across modalities, reinforcing the need for imbalance-aware training in VRU detection.  


\subsection{Augmentation Results}

To evaluate the impact of data augmentation on model robustness, experiments were conducted using a subset of 4,000 RGB images with an adjusted class-weighted loss function, an input size of 640 pixels, a batch size of four, and six frozen backbone layers. Three training configurations were compared: no augmentation, light augmentation, and heavy augmentation. The results are summarised in Table~\ref{Augmentation Experiment}.

\begin{table}[H]
    \centering
    \begin{tabular}{|p{3.4em}|c|c|c|c|}
    \hline
        \textbf{Level} & \textbf{Box} & \textbf{Recall} & \textbf{mAP50} & \textbf{mAP50:90}  \\
        \hline
        \hline
        None & 0.646 & 0.480 & 0.531 & 0.316 \\
        \hline
        Light & 0.613 & 0.423 & 0.463 & 0.273 \\
        \hline
        Heavy & 0.581 & 0.389 & 0.412 & 0.237 \\
        \hline
    \end{tabular}
    \caption{\small Augmentation Experiment Results}
    \label{Augmentation Experiment}
\end{table}

As seen in Table~\ref{Augmentation Experiment}, both the light and heavy enhancement strategies led to reductions in mAP compared to baseline without enhancement. Specifically, the heavy-augmentation regime resulted in the lowest performance, with mAP50 dropping to 0.412 and recall decreasing to 0.389. Light augmentation produced a moderate performance reduction (mAP50 = 0.463, recall = 0.423), but the decline was less severe than in the heavily augmented case.  

\begin{figure}[h]
    \centering
    \includegraphics[width=0.8\linewidth]{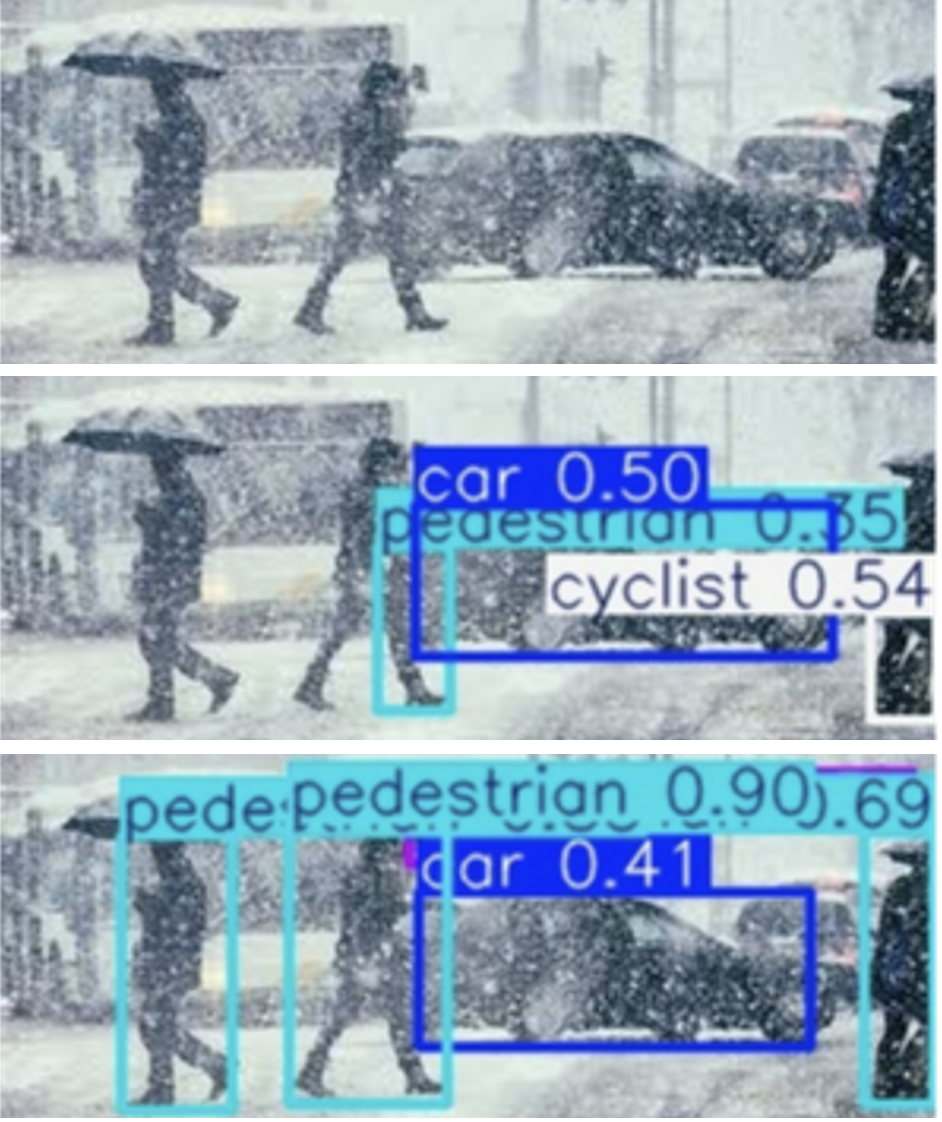}
    \caption{Comparison of model detection performance in adverse weather: original image, results without augmentation, and improved detections with augmentation, highlighting enhanced robustness.}
    \label{augmentations image}
\end{figure}

Despite the observed degradation in standard metrics, the qualitative evaluation illustrated the benefits of increase under adverse conditions. Figure~\ref{augmentations image} shows that the light-augmented model successfully detected pedestrians obscured by snow, whereas the non-augmented model failed to do so. This highlights that augmentation improves model generalisation to unseen and challenging scenarios, even if it slightly reduces benchmark performance.  

In addition, the thermal data set was expanded using an RGB-to-thermal image translation technique \cite{lee2023edge}. This approach enriched the thermal training set with synthetic samples, enabling the model to better detect VRUs under limited visibility or when genuine thermal data were sparse.  

\subsection{Final Training Results}

Based on the findings of the preceding experiments, the final training phase focused on both RGB and thermal models with 7-Class filtration. To further address class imbalance, additional filtering was applied to emphasise minority VRU categories (cyclist, motorcycle, scooter). The selected training parameters included an input resolution of 640 pixels, a batch size of four, six frozen backbone layers, and 100 training epochs with a patience setting of 50.  

In addition to the baseline RGB and thermal models, three additional models were trained: (i) an RGB model with light augmentations, (ii) a thermal model with light augmentations, and (iii) a thermal model incorporating an RGB-to-thermal image conversion to expand the dataset with synthetic thermal samples. This configuration yielded five final models for comparison. To ensure fairness, the two RGB models shared a single validation set, while the three thermal models were validated against a separate but consistent thermal validation set. Importantly, only non-augmented images were used for validation.  

The RGB-to-thermal image conversion process, used to enrich the thermal training dataset, is illustrated in Figure~\ref{RGB to TIR Image Conversion}. Although a degradation in image quality is evident, such transformations may contribute positively to model robustness by exposing the detector to more diverse representations of VRUs.  

Table~\ref{Final Training Results} presents the performance of the final models, while Table~\ref{Class instances for final training models} summarises the class distributions between training and validation splits.  

\begin{table}[h]
    \centering
    \begin{tabular}{|p{4.5em}|p{4em}|p{4em}|p{3em}|p{3em}|}
    \hline
        \textbf{Model} & \textbf{mAP50} & \textbf{mAP 50:90} & \textbf{Box} & \textbf{Recall} \\
        \hline
        \hline
        RGB & 0.620 & 0.374 & 0.713 & 0.549 \\
        \hline
        RGB Aug & 0.606 & 0.442 & 0.780 & 0.638 \\
        \hline
        Thermal & 0.724 & 0.442 & 0.780 & 0.638 \\
        \hline
        Thermal Aug & 0.632 & 0.398 & 0.749 & 0.512 \\
        \hline
        Thermal RGB-TIR & 0.696 & 0.452 & 0.618 & 0.689 \\
        \hline
    \end{tabular}
    \caption{\small Final training results}
    \label{Final Training Results}
\end{table}

\begin{table*}[tb]
    \centering
    \begin{tabular}{|p{5.3em}|p{5.6em}|p{4.6em}|p{7.1em}|p{6.4em}|p{12em}|}
    \hline
        \textbf{Class} & \textbf{RGB Train} & \textbf{RGB Val} & \textbf{Therm Train} & \textbf{Therm Val} & \textbf{Therm RGB-TIR Train}  \\
        \hline
        \hline
         Car & 89639 & 9814 & 31615 & 1107 & 84772 \\
         \hline
         Pedestrian & 40320 & 4437 & 18985 & 980 & 44019 \\
         \hline
         Cyclist & 14207 & 1200 & 7237 & 170 & 13418 \\
         \hline
         Truck & 3541 & 465 & 240 & 9 & 3214 \\
         \hline
         Bus & 2170 & 266 & 675 & 28 & 2055 \\
         \hline
         Motor cycle & 4663 & 529 & 1116 & 55 & 3651 \\
         \hline
         Scooter & 38 & 0 & 15 & 0 & 15 \\
         \hline
    \end{tabular}
    \caption{\small Class instances for final training models}
    \label{Class instances for final training models}
\end{table*}

As expected, the inclusion of augmentations led to slight reductions in standard performance metrics. The RGB augmented model showed only a marginal drop in mAP50 (1.4\%), while the thermal augmented model experienced a larger decline of 9.2\%. However, these decreases are acceptable given that the improvements improve robustness in adverse weather and lighting conditions, where detection reliability is often more critical than the precision of the benchmark.  

The RGB-to-thermal model produced competitive results, with an mAP50 of 0.696 and a strong recall (0.689), suggesting that synthetic thermal data can enhance performance, particularly when real thermal data are limited. In general, the thermal baseline model achieved the highest precision (mAP50 = 0.724), but the augmented and converted models demonstrated better resilience, supporting the use of diverse training strategies for robust VRU detection.

\section{Conclusions}

This study presented a multimodal detection framework for vulnerable road users (VRUs), integrating RGB and thermal infrared imaging with a fine-tuned YOLOv8 model. By leveraging a composite dataset drawn from KITTI, BDD100K, and Teledyne FLIR, and applying strategies such as backbone freezing, class re-weighting, and augmentation, the system was optimised for both accuracy and efficiency on edge devices. Experimental results demonstrated that a 640-pixel input resolution and partial backbone freezing achieved the best trade-off between detection performance and computational cost. Class-weighted loss functions substantially improved recall for minority VRU classes, while light augmentations improved robustness under adverse weather conditions.

The final evaluations confirmed the complementary strengths of the RGB and thermal models, with thermal achieving the highest precision and RGB to thermal enhancement improving recall. Together, these findings highlight the potential of multimodal, edge-based detection systems to improve VRU safety in real-world traffic environments, particularly at intersections where collision risks are greatest.

Future work will extend this approach through larger and more diverse datasets, real-world field deployments, and the integration of additional sensing modalities such as radar or LiDAR. In addition, the incorporation of temporal tracking and trajectory prediction could further enhance the system’s ability to anticipate and prevent VRU–vehicle conflicts.


\bibliographystyle{ieeetr}
\bibliography{acra}

\end{document}